\documentclass[twocolumn]{article}
\usepackage[pdftex]{graphicx}
\usepackage{authblk,color}
\usepackage{url}
\usepackage{amsmath}
\usepackage{tabu}
\usepackage{booktabs}

\title{
The Impact of Prompts on Zero-Shot Detection of\\ AI-Generated Text
}

\date{\empty}
\author{Kaito Taguchi\thanks{Email:taguchi.kaito.834@s.kyushu-u.ac.jp}}
\author{Yujie Gu\thanks{Email:gu@inf.kyushu-u.ac.jp}}
\author{Kouichi Sakurai\thanks{Email:sakurai@inf.kyushu-u.ac.jp}}

\affil{Kyushu University, Fukuoka, Japan\\
}

\begin{document}
\twocolumn[
  \begin{@twocolumnfalse}
    \maketitle
    \begin{abstract}
    In recent years, there have been significant advancements in the development of Large Language Models (LLMs). While their practical applications are now widespread, their potential for misuse, such as generating fake news and committing plagiarism, has posed significant concerns. 
    To address this issue, 
    detectors have been developed to evaluate whether a given text is human-generated or AI-generated. 
    Among others,
    zero-shot detectors stand out as effective approaches that do not require additional training data and are often likelihood-based.
    In chat-based applications, users commonly input prompts and utilize the AI-generated texts. 
    However, zero-shot detectors typically analyze these texts in isolation, neglecting the impact of the original prompts. 
    It is conceivable that this approach may lead to a discrepancy in likelihood assessments between the text generation phase and the detection phase.
    So far, there remains an unverified gap concerning how the presence or absence of prompts impacts detection accuracy for zero-shot detectors. 
    In this paper, we introduce an evaluative framework to empirically analyze the impact of prompts on the detection accuracy of AI-generated text. 
    We assess various zero-shot detectors using both white-box detection, which leverages the prompt, and black-box detection, which operates without prompt information. 
    Our experiments reveal the significant influence of prompts on detection accuracy. Remarkably,
    compared with 
    black-box detection without prompts, the white-box methods using prompts demonstrate an increase in AUC of at least $0.1$ across all zero-shot detectors tested. Code is available: \url{https://github.com/kaito25atugich/Detector}.\\ \\ 

    \end{abstract}
  \end{@twocolumnfalse}
  ]

\section{Introduction}
Recent years have seen significant advancements in the development of Large Language Models (LLMs)~\cite{openai2023,microsoft2023,google2023}, and their practical applications have become widespread. Meanwhile, their potential misuse have raised significant concerns. For example, the generation of fake news and plagiarism using LLMs is a notable issue. Detectors that evaluate whether a given text is human-generated or AI-generated serve as a defense mechanism against such misuse. 

Detectors for AI-generated text can be broadly classified into three categories: a zero-shot detector leveraging statistical properties~\cite{gehrmann2019,mitchell2023,bao2023,su2023,hans2024,liu2024,sasse2024,tulchinskii2023}, a detector employing supervised learning~\cite{solaiman2019,hu2023,dou2024,liuz2023}, and a detector utilizing watermarking~\cite{kirchenbauer2023,ren2023}.

\begin{table*}[h]
\centering
\caption{Summary of Zero-short Detectors}
\label{tab:sum_detectors}
{%
\begin{tabular}{l l}
\hline
Method & Summary \\ \hline \\

Log-likelihood & \begin{tabular}{l} Detect using the log likelihood of the given text. \end{tabular}  \\ \\ 

Rank & \begin{tabular}{l} Calculate the likelihood of the given text and convert the likelihood of each \\ token into ranks based on the entire vocabulary, then use this to detect. \end{tabular} \\ \\
Log-Rank & \begin{tabular}{l} Calculate the likelihood of the given text and transform the likelihood of each\\ token  into ranks based on the entire vocabulary, then apply logarithm to these\\ ranks for detection. \end{tabular}  \\ \\
Entropy &  \begin{tabular}{l} Detect by calculating entropy using the likelihood of tokens in the vocabulary. \end{tabular} \\ \\
DetectGPT~\cite{mitchell2023} & \begin{tabular}{l} Using a masked language model, randomly replace words in the text. Observe\\ the likelihood of the replaced text and the original text using a scoring model, \\ and utilize the change to detect alterations. \end{tabular} \\ \\ 
 
FastDetectGPT~\cite{bao2023} & \begin{tabular}{l} Replace the mask model in DetectGPT with a auto-regressive model similar to \\ the scoring model. Sample words randomly from the vocabulary to replace words.\\ Calculate scores in the same manner as DetectGPT. \end{tabular}  \\ \\
LRR~\cite{su2023} &  \begin{tabular}{l} Detect using the ratio of log-likelihood to log-rank. \end{tabular} \\ \\

NPR~\cite{su2023} &  \begin{tabular}{l} Similar to DetectGPT, utilize logarithmic ranks rather than logarithmic\\ likelihood  for scoring calculation. \end{tabular} \\\\

Binoculars~\cite{hans2024} &  \begin{tabular}{l} Utilize models trained with slightly different amounts of data and calculate the \\ perplexity  of each model. Then leverage the difference in perplexity for detection. \end{tabular} \\ \\
\hline
\end{tabular}}
\end{table*}

Zero-shot detectors, such as DetectGPT~\cite{mitchell2023}, which do not require additional training, are designed in many methods using likelihood-based scores. 
A summary of zero-shot detectors is illustrated in Table \ref{tab:sum_detectors}. In other words, the zero-shot detection is carried out by replicating the likelihood at the generation phase. 
When using LLMs, we usually input prompts and utilize the generated output. However, at the detection phase, it is anticipated that reproducing likelihood becomes challenging due to the absence of the contextual information provided by prompts. 
It may potentially result in differences in likelihood evaluations between the text generation and detection stages. 


In this paper, we assess to what extent this phenomenon affects likelihood-based zero-shot detectors.
The contributions of this study are as follows:

\begin{itemize}
\item We propose two methods for detecting AI-generated text using zero-shot detectors: white-box detection, which leverages the prompts used to generate the text, and black-box detection, which detects AI-generated text without relying on a prompt.
\item Extensive experiments demonstrate a decrease in detection accuracy for existing zero-shot detectors in black-box detection.
\item Indication of the significance of sample size and its ratio for the robustness of the Fast series detectors.
\end{itemize}

\section{Related work}
In the context of intentionally undermining detection accuracy using prompts, two main categories of studies can be identified. The first category involves the deliberate crafting of prompts with malicious intent to deliberately reduce detection accuracy. In contrast, the second category encompasses research that employs tasks with benign prompts, devoid of malicious intent.

\subsection{Malicious prompts}

First, we delve into studies that specifically concentrate on the deliberate creation of malicious prompts.

In \cite{koike2023}, Koike et al. proposed OUTFOX, utilizing in-context learning with the problem statement $P$, human-generated text $H$, and AI-generated text $A$. By constructing prompts such as ``$p_i \in P\rightarrow h_i \in H$ is the correct label by humans, and $p_i \in P\rightarrow a_i \in A$ is the correct label by AI," they aim to generate text for a given problem statement in such a way that the generated text aligns with human-authored content. This approach makes the detection of artificially generated content challenging.

Shi et al. conducted an attack on OpenAI's Detector~\cite{openai_text_classifier2023} by employing an Instructional Prompt, confirming a decrease in detection accuracy~\cite{shi2023}. The Instructional Prompt involves adding a reference text $X_{ref}$ and an instructional text $X_{ins}$ with characteristics that reduce the detection accuracy to the original input $X$, thereby undermining the detection accuracy.

In~\cite{lun2023}, Lu et al. proposed SICO, a method that lowers detection accuracy by instructing the model within prompts to mimic the writing style of human-authored text and updating the content of the instructions to reduce detection accuracy.

Kumarage et al. proposed an attack named Soft Prompt, which generates a vector using reinforcement learning to induce misclassification by detectors. This Soft Prompt vector is then used as input for the DetectGPT and RoBERTa-base detectors~\cite{solaiman2019}, demonstrating a decrease in detection accuracy~\cite{kumarage2023}.

\subsection{Benign prompts}

We review cases involving tasks with benign prompts here.

Liu et al. conducted experiments using the CheckGPT model, an approach based on supervised learning. Their findings indicate that when using different prompts, although all surpass 90\%, there is an experimental demonstration of approximately a 7\% decrease in detection accuracy~\cite{liuz2023}.

Dou et al.~\cite{dou2024} performed experiments envisioning the utilization of LLMs by students. In their study, they demonstrated a decrease in DetectGPT's detection accuracy when prompts were employed.

Hans et al.~\cite{hans2024} pointed out the difficulty in reproducing likelihoods depending on the presence or absence of prompts, using unique prompts like ``Write about a capybara astronomer." In response to the capybara problem, they proposed Binoculars.

We assume performing benign tasks such as summarization. Therefore, unlike malicious prompt attacks, there is no need to deliberately choose prompts that would lower accuracy using the detector when constructing prompts, nor is there a requirement to collect pairs of data for in-context learning.

On the other hand, Dou et al.~\cite{dou2024} experimentally demonstrated unintended decreases in detection accuracy. However, they did not delve into why the accuracy decreases or make references to other likelihood-based zero-shot detectors. Additionally, Hans et al.~\cite{hans2024} did not provide specific verification regarding the impact of a detector knowing or not knowing the prompt on detection accuracy. Therefore, the resilience of Binoculars to changes in likelihood due to prompts has not been adequately assessed. The supervised learning based approach~\cite{liuz2023} is excluded from our experiments in this context.

In this study, we demonstrate that even in ordinary tasks such as summarization, the presence or absence of prompts unintentionally leads to a decrease in accuracy when using likelihood-based zero-shot detectors.

\section{Preliminary}
\subsection{Language model}
A model that captures the probability of generating words or sentences is referred to as a language model. 
Let $V$ represent the vocabulary.
The language model for a word sequence of length $n$, denoted as $x_1, x_2, \ldots, x_n$ where $x_i \in V$, is defined by the following \eqref{eq:language model}. 
\begin{equation}
\label{eq:language model}
    p(x_1, x_2, ..., x_n )=\prod_{t=1}^n p(x_t|x_1, ..., x_{t-1})
\end{equation}

\subsection{Existing zero-shot detectors}
We provide a brief introduction to existing zero-shot detectors, summarized in Table 1. Here, $P_{T_\theta}$ refers to the language model utilized for detection. The vocabulary $V$ is composed of $C$ tokens. The input text $S$ is composed of $N$ tokens, represented as $S=\{S_1, S_2, ..., S_N\}$, and the token sequence from $S_1$ to $S_{i-1}$ is denoted as $S_{<i}$. 

\subsubsection{Log-Likelihood}
The log-likelihood is a method that utilizes the likelihood of tokens composing a text for detection. The formula is presented in \eqref{eq:loglikelihood}. The log-likelihood is the average of the log-likelihoods of tokens constituting a given text.
\begin{equation}
\label{eq:loglikelihood}
\text{Log-likelihood}=\frac{1}{N-1} \sum_{i=2}^N \log P_{T_\theta } (S_i |S_{<i}).
\end{equation}

\subsubsection{Entropy}
Entropy is a method that utilizes the entropy of the vocabulary for detection. The formula is shown in  \eqref{eq:entropy}. Entropy is calculated using the likelihood of the vocabulary, taking the average across each context.
\begin{align}
\label{eq:entropy}
\text{Entropy}=\frac{-1}{N-1}\sum_{i=2}^N \sum_{j=1}^C P_{T_\theta} (j|S_{<i})\log P_{T_\theta} (j|S_{<i}).
\end{align}

\subsubsection{Rank}
Rank is a method that utilizes the order of likelihood magnitude of tokens in the vocabulary when sorted. The formula is presented in \eqref{eq:rank}. Rank is the average position of tokens constituting a given text. The function $sort$ is a function that sorts the given array in descending order, and $index$ is a function that, given an array and an element as input, returns the index of the element within the given array.

\begin{equation}
\label{eq:rank}
\text{rank}=\frac{-1}{N-1} \sum_{i=2}^N index(sort(\log P_{T_\theta} (S_i|S_{<i})),S_i).
\end{equation}

\subsubsection{DetectGPT}
The language model aims to maximize likelihood during text generation, whereas humans create text independently of likelihood. DetectGPT focuses on this phenomenon and posits a hypothesis that by rewriting certain words, the likelihood of the text decreases for AI-generated content and can either increase or decrease for human-generated content~\cite{mitchell2023}.

\begin{figure*}[tb]
    \centering
    \includegraphics[width=11cm]{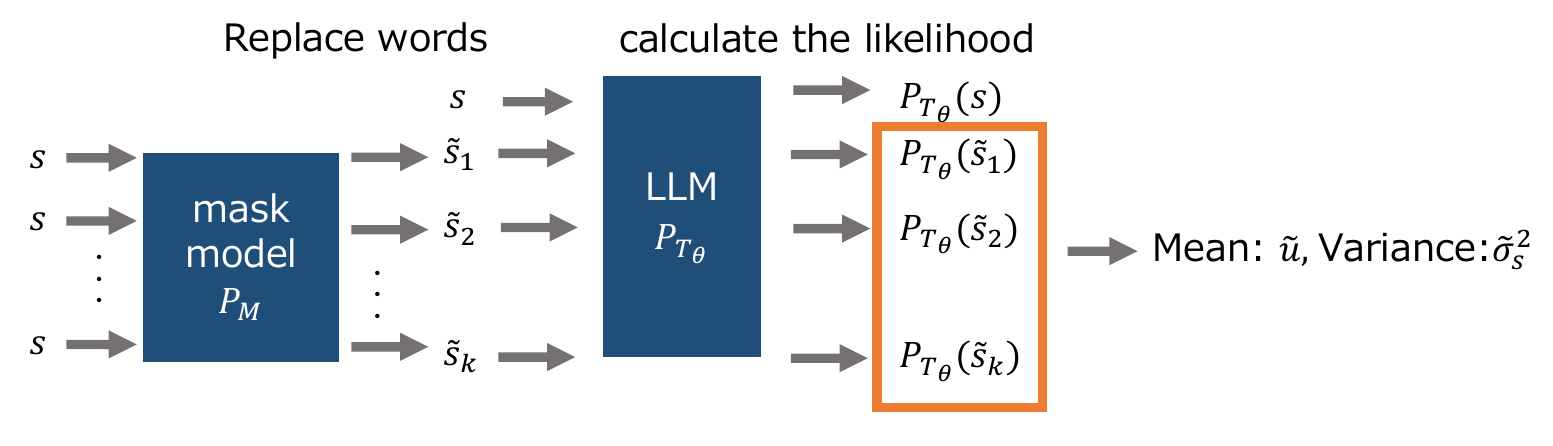}
    \caption{DetectGPT Overview}
    \label{fig:detectGPT}
\end{figure*}

The overview of DetectGPT is presented in Figure \ref{fig:detectGPT}. The replacement process is achieved by utilizing a mask model$P_M$, such as T5~\cite{raffel2020}, on some of the words contained in the given text $S$. This operation is repeated for a total of $k$ iterations, and the average log-likelihood of the obtained $k$ replacement texts is then computed.  \eqref{eq:perturbation_discrepancy} represents the score, calculating the difference between the log-likelihood of the original text and the average log-likelihood of the acquired replacement texts. It is permissible to standardize by dividing by the standard deviation of the log-likelihood of the replacement texts. If the score is above the threshold $\varepsilon$, it is deemed to be AI-generated text.

\begin{equation}
\label{eq:perturbation_discrepancy}
\text{DetectGPT} = \frac{\log P_{T_\theta}(S)-\tilde{m}}{\tilde{\sigma_S}}
\end{equation}
where
\begin{align*}
    \tilde{m} &= \frac{1}{k}\sum^k_{i=1}\log P_{T_\theta}(\tilde{S_i})\\
    \tilde{\sigma_S}&=\frac{1}{k-1}\sum^k_{i=1}(\log P_{T_\theta}(\tilde{S_i})-\tilde{u})^2
\end{align*}
and $\tilde{S_i}\sim P_M(S_i)$ represent the mean, sample variance, and a sample from $P_M(S_i)$, respectively.

\subsubsection{FastDetectGPT}
In~\cite{bao2023}, Bao et al. highlighted challenges in DetectGPT's use of different models for substitution and score calculation, as well as the cost-related aspect of requiring model access for each substitution iteration. In response, FastDetectGPT is a modified detector that reduces access to the model, addressing the cost issue while enabling substitutions. Although the methodology involves setting hypotheses similar to DetectGPT, there is no fundamental change. It still operates on the assumption that ``AI-generated text is likely to be around the maximum likelihood, whereas human-generated text is not."

\begin{figure*}[tb]
    \centering
    \includegraphics[width=11cm]{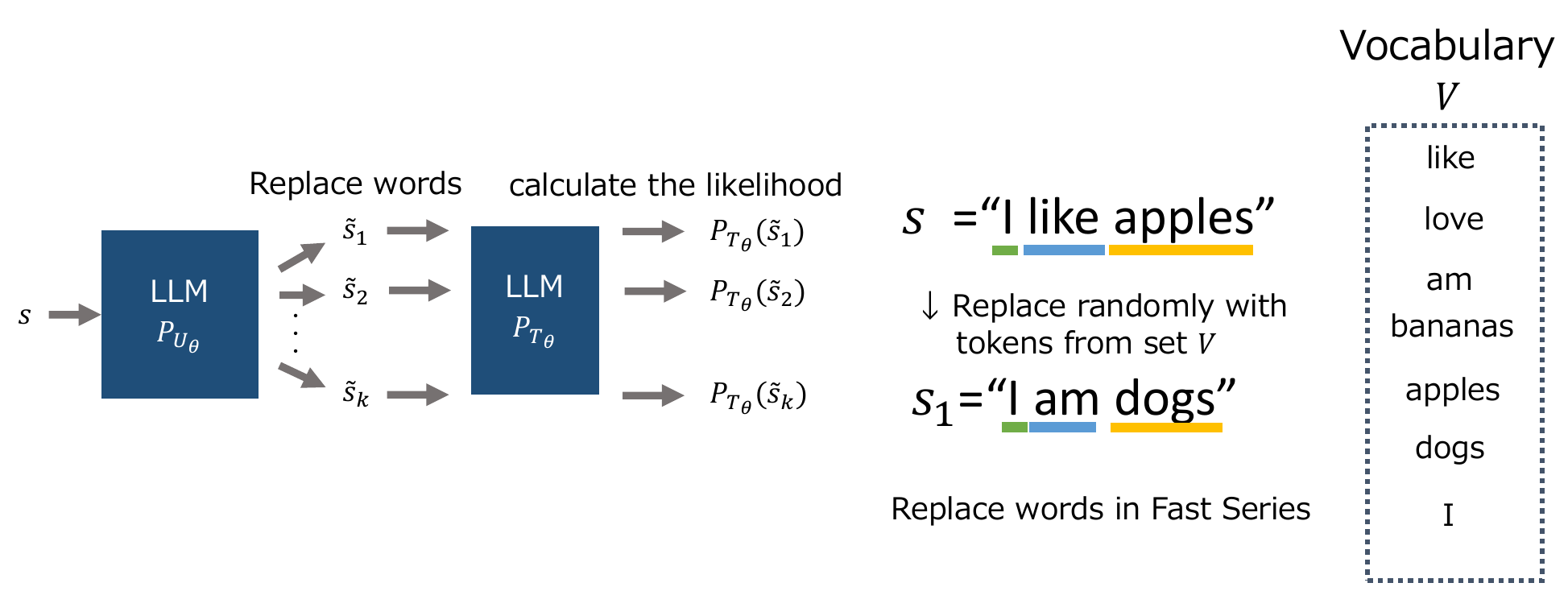}
    \caption{FastDetectGPT and Sampling Overview}
    \label{fig:fastdetectgpt}
\end{figure*}

We present the overall architecture of FastDetectGPT in Figure \ref{fig:fastdetectgpt}. In FastDetectGPT, the substitution process is replaced with an alternative method that does not rely on a mask model. Similar to the detection model, it utilizes an autoregressive model, and $P_{T_\theta}$ and $P_{U_\theta}$ can be the same. The substitution for the $i$-th word involves randomly extracting a word from the next-word list, considering the context up to the $(i-1)$-th word in the input text, and replacing the word with the chosen one. In other words, performing this substitution $N$ times results in the substituted text $\tilde{S}$, and by conducting sampling during word selection, the replacement process generates $k$ substitution texts in a single access.

The subsequent score calculation is omitted as it follows the same procedure as DetectGPT.

\subsubsection{LLR \& NPR}
LLR (\textbf{L}ikelihood \textbf{L}og-Rank \textbf{r}atio) and NPR (\textbf{N}ormalized \textbf{p}erturbed log \textbf{r}ank) are classical log-rank enhancement techniques proposed by Su et al.~\cite{su2023}. Both methods have simple configurations. LLR literally takes the ratio of log-likelihood to log-rank, as expressed in  \eqref{eq:llr}. Here, $r_\theta$ represents the rank when using $P_{T_\theta}$.

\begin{equation}
\label{eq:llr}
LRR = -\frac{\sum^t_{i=1}\log P_{T_\theta}(S_i|S_{<i})}{\sum^t_{i=1}\log r_\theta(S_i|S_{<i})}
\end{equation}

On the other hand, NPR, like DetectGPT, performs the substitution of words in the text $k$ times. It takes the ratio of the average log-rank of the obtained substituted texts to the log-rank of the original text. This is defined in  \eqref{eq:npr}.

\begin{equation}
\label{eq:npr}
NPR = \frac{\frac{1}{k}\sum^k_{p=1}\log r_\theta(\tilde{S}_p)}{\log r_\theta(S)}
\end{equation}

\subsubsection{Binoculars}

\begin{figure*}[tb]
    \centering
    \includegraphics[width=11cm]{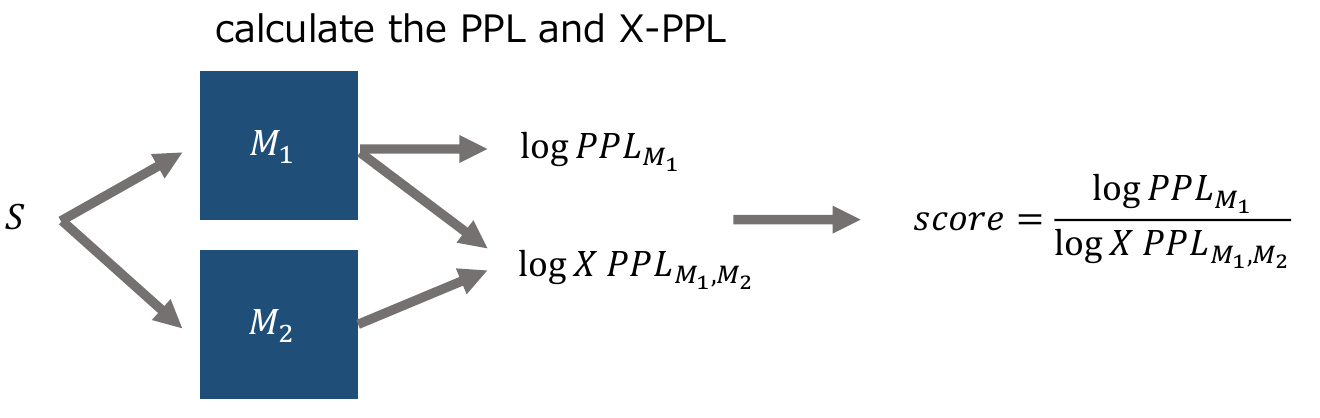}
    \caption{Binoculars Overview}
    \label{fig:binoculars}
\end{figure*}

Hans et al. proposed Binoculars, a detection method utilizing two closely related language models, Falcon-7b~\cite{almazrouei2023} and Falcon-7b-instruct, by employing a metric called cross-perplexity~\cite{hans2024}. The overall framework is illustrated in Figure \ref{fig:binoculars}.

Let the first model be denoted as $M_1$ (such as Falcon-7b), and the second model as $M_2$ (like Falcon-7b-instruct). In this case, using $M_1$, we calculate the log perplexity as shown in  \eqref{eq:logperplexity}.
\begin{equation}
\label{eq:logperplexity}
\log PPL_{M_1}(S)=-\frac{1}{N}\sum^N_{i=1}\log(M_1(S_i|S_{<i}))
\end{equation}

Next, using $M_1$ and $M_2$, we calculate the cross-perplexity, as shown in \eqref{eq:crossperplexity}. Here, the symbol $\cdot$ represents the dot product.

\begin{multline}
\label{eq:crossperplexity}
\log X \mathchar`- PPL_{M_1, M_2}(S)= \\ -\frac{1}{N}\sum^N_{i=1}\sum_{j=1}^C M_1(j|S_{<i})\cdot\log(M_2(j|S_{<i}))
\end{multline}

The score in Binoculars is determined by  \eqref{eq:scorebinoculars}.

\begin{equation}
\label{eq:scorebinoculars}
B_{M_1, M_2}(S)=\frac{\log PPL_{M_1}(S)}{\log X\mathchar`-PPL_{M_1, M_2}(S)}
\end{equation}

\section{Proposal}
In this study, we propose a detection flow to investigate the impact of prompts on likelihood.

Before presenting the experimental setup, we introduce an additional detection method.

\subsection{FastNPR}
Word replacements in NPR are performed using a masked model. In this research, aiming for cost reduction, we also employ FastNPR, a method that replaces word replacements with sampling, akin to FastDetectGPT.

\subsection{Detection methods}
\begin{figure*}[tb]
    \centering
    \includegraphics[width=11cm]{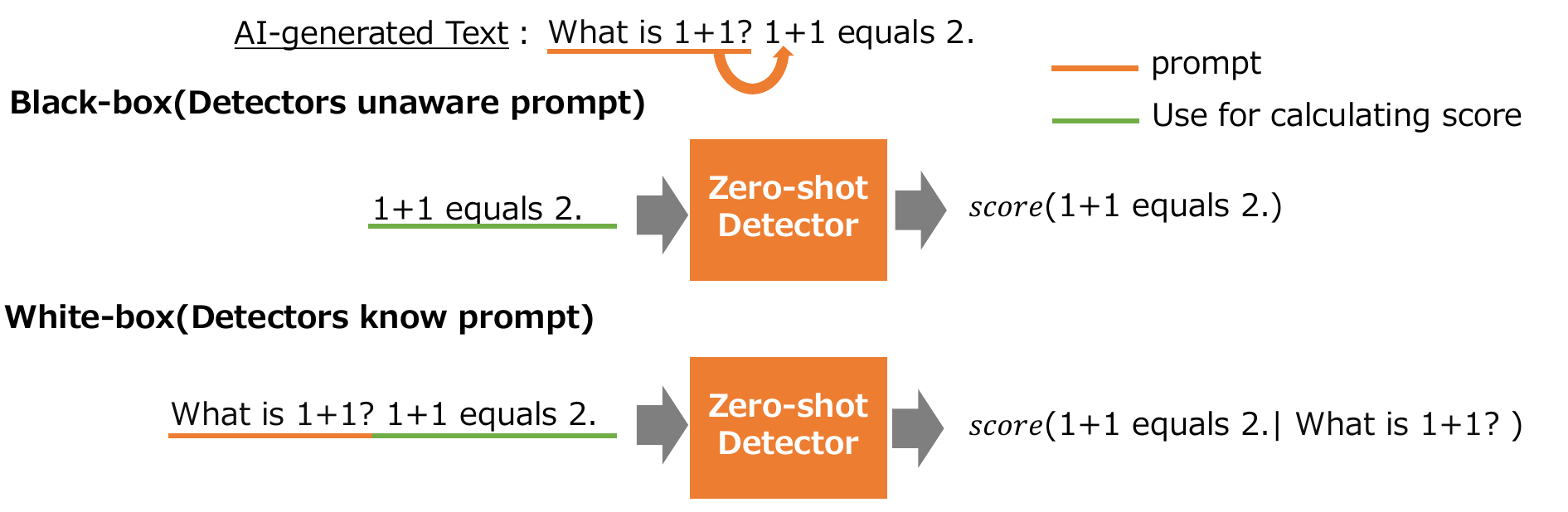}
    \caption{Proposed Detection Methods Overview}
    \label{fig:proposal}
\end{figure*}

We explain the detection methodology. For the purpose of the explanation, let $x$ represent the text to be detected, and if $x$ is an AI-generated text, let $p$ denote the prompt used for its generation. Detection can be categorized into two patterns: Black-box detection and White-box detection. An overview is presented in Figure \ref{fig:proposal}.

Black-box detection occurs when the detector is unaware of prompt information, essentially mirroring existing detection methods. In this scenario, only the content of $x$ is provided to the detector.

White-box detection, on the other hand, involves the detector having knowledge of prompt information. For human-generated text, only $x$ is input. In the case of AI-generated text, the input consists of $p+x$. It is important to note that, in White-box detection, the prompt is used solely for likelihood calculation and is not included in the score computation. 


\section{Experiment}
\subsection{Configuration}
To begin, we utilize the GPT2-XL~\cite{radford2019} as the detection model, excluding Binoculars. Due to GPU constraints, Binoculars employs the pre-trained and instruct-tuned Phi1.5~\cite{liy2023} instead of Falcon.

For DetectGPT and NPR, we generate five replacement sentences for 10\% of the entire text, while the Fast series generates 10,000 replacement sentences. T5-Large~\cite{raffel2020} is used for word replacement in DetectGPT and NPR, while the Fast series employs the GPT2-XL, the same detection model.

Also, we use the XSum dataset~\cite{narayan2018}. For human-generated text, we extract 200 samples from the XSum dataset, and for AI-generated text, we employ the Llama2 7B Chat model~\cite{touvron2023}, generating up to 200 tokens. The prompt used is ``Would you summarize the following sentences, please? {text}".

\subsection{Result}

\begin{table}
\centering
\caption{Detection of Generated Summaries: Discrepancies Between Cases with and Without Prompts}
\label{tab:sum_detection}
{%
\begin{tabular}{lllllll}
\hline
Method & Black-box & White-box \\ \hline
DetectGPT & 0.453 & 1.000 \\
FastDetectGPT & 0.819 & 0.958 \\
LRR & 0.532 & 0.995 \\
NPR & 0.560 & 0.934 \\
FastNPR & 0.768 & 0.993 \\
Entropy & 0.330 & 0.978 \\
Log-likelihood & 0.474 & 0.998 \\
Rank & 0.432 & 0.977 \\
Log-Rank & 0.485 & 0.999 \\
Binoculars & 0.877 & 0.999 \\
\hline
\end{tabular}}
\end{table}

As evident from the results in Table \ref{tab:sum_detection}, white-box detection exhibits higher accuracy, while black-box detection shows lower accuracy. As anticipated, modifying likelihood through prompts leads to a decrease in the detection accuracy of likelihood-based detectors. Notably, there is a consistent decrease of 0.1 or more across all methods, highlighting a significant observation.

Binoculars and the Fast series detectors demonstrate robustness compared to other methods. In particular, the Fast series detector maintains the same scoring calculation as conventional methods, suggesting robustness factors in the sampling process. For further verification, we conduct additional experiments.

In this experiment, we investigate the differences in detection accuracy when varying the replacement ratio, indicating the extent to which tokens in the text are replaced, and the sample size, representing the number of replacement sentences. DetectGPT and NPR require the use of a masked language model to replace plausible tokens, making replacement not always feasible, especially for higher replacement percentages. Therefore, we primarily vary the replacement ratio in the Fast series to conduct the investigation.

The results for DetectGPT are presented in Table \ref{tab:anal_detectgpt}, and the results for NPR are shown in Table \ref{tab:anal_npr}.
From these results, it is evident that increasing the replacement ratio and sample size helps mitigate the decrease in detection accuracy. This observation is similar to Chakraborty et al.'s assertion that increasing the sample size can enable detection if the distribution slightly differs~\cite{chakraborty2023}. 

However, in our validation, the improvement in accuracy plateaus at around 10 samples, reaching a maximum AUC of approximately 0.8, which is not considered high. Particularly in recent years, there is a trend toward practical applications, emphasizing high true positive rates at low false positive rates, suggesting that at least an AUC in the late 0.9s would be necessary \cite{krishna2023,hans2024}. Furthermore, the lack of improvement in detection accuracy with DetectGPT and NPR may be attributed to the limited number of substitutable tokens.

\begin{table}[ht]
\centering
\caption{Effect of Substitution Rate(SR) and Sample Size(SS) Variation on AUC(DetectGPT)}
\begin{tabular}{llll}
\hline
Method & SR & SS & AUC \\
\hline \hline
FastDetectGPT & 10\% & 5 & 0.640 \\
FastDetectGPT & 20\% & 5 & 0.697 \\
FastDetectGPT & 100\% & 5 & 0.779 \\
FastDetectGPT & 10\% & 10 & 0.704 \\
FastDetectGPT & 20\% & 10 & 0.739 \\
FastDetectGPT & 100\% & 10 & 0.821 \\
FastDetectGPT & 100\% & 10000 & 0.819 \\
DetectGPT & 10\% & 5 & 0.453 \\
DetectGPT & 20\% & 5 & 0.522 \\
DetectGPT & 30\% & 5 & 0.490 \\
DetectGPT & 10\% & 10 & 0.446 \\
DetectGPT & 30\% & 10 & 0.446 \\
\hline
\end{tabular}

\label{tab:anal_detectgpt}
\end{table}
\begin{table}[ht]
\centering
\caption{Effect of Substitution Rate(SR) and Sample Size(SS) Variation on AUC(NPR)}

\begin{tabular}{llll}
\hline
Method & SR & SS & AUC \\
\hline \hline
FastNPR & 10\% & 5 & 0.628 \\
FastNPR & 20\% & 5 & 0.661 \\
FastNPR & 100\% & 5 & 0.747 \\
FastNPR & 10\% & 10 & 0.647 \\
FastNPR & 20\% & 10 & 0.715 \\
FastNPR & 100\% & 10 & 0.750 \\
FastNPR & 100\% & 10000 & 0.763 \\
NPR & 10\% & 5 & 0.560 \\
NPR & 20\% & 5 & 0.590 \\
NPR & 30\% & 5 & 0.577 \\
NPR & 10\% & 10 & 0.589 \\
NPR & 30\% & 10 & 0.588 \\
\hline
\end{tabular}

\label{tab:anal_npr}
\end{table}

\section{Discussions}
\subsection{Hypotheses for zero-shot detectors}
While our investigation has focused solely on prompts, similar phenomena could potentially be observed with other elements. For instance, variations in Temperature or Penalty Repetition between the generation and detection stages might introduce differences in the selected tokens, making detection challenging based on likelihood. Generalizing these observations, we hypothesize that any act that fails to replicate the likelihood during language generation could undermine the detection accuracy of Zero-shot detectors relying on likelihood from next-word prediction.

\subsection{Common tasks}
While our investigation has focused on summary text generation, there are several other potential tasks to consider, such as paraphrase generation, story generation, and translation text generation. It is plausible that detection accuracy could also decrease in these common tasks. Since these tasks may be utilized without malicious intent, it is crucial to conduct similar evaluations for them.

\subsection{Relevance to paraphrase attacks}
Paraphrase generation, as briefly discussed in the previous section, assumes a single act. However, currently known paraphrase attacks \cite{sadasivan2023,krishna2023,shi2023,hu2023} involve generating paraphrases for each sentence and combining the results. While paraphrase attacks using masked language models may have a slightly different structure, as they utilize both preceding and succeeding contexts for word replacement, it can be argued that reproducing likelihood during detection becomes challenging. Therefore, paraphrase attacks can be viewed as more complex versions of the tasks verified in this study.

\subsection{Text length} 
In the current experiment, the generated texts were fixed at 200 tokens. The length of tokens may impact the ease of reproducing likelihood. Therefore, it would be beneficial to conduct further verification with longer texts. Tasks such as narrative generation, where the length of the text is not a concern, may be suitable for such investigations.

\subsection{Number of parameters}
In this study, each detection method utilized a language model of approximately 1 billion parameters. It would be of interest to investigate whether increased robustness can be observed when experimenting with larger language models. Conversely, there are experimental studies that have demonstrated the ability of smaller language models to achieve a higher likelihood for AI-generated texts across a broader range of language models \cite{mireshghallah2023}. Considering these findings, conducting experiments with smaller language models and verifying if there are differences in robustness could also provide valuable insights.

\subsection{Relationship with supervised learning detectors}

Even when using supervised learning, it has been noted that generated text from prompt-based tasks may exhibit decreased detection accuracy~\cite{liuz2023}. However, there is a possibility that these models could be more robust compared to zero-shot detectors. For instance, RADAR \cite{hu2023} achieved an AUC of 0.939 in the task used in this experiment. In comparison, the RoBERTa-large detector \cite{solaiman2019} had an AUC of 0.767. This suggests that robust detectors against paraphrase attacks might demonstrate similarly robust results in other tasks.

\subsection{Relationship with watermarking}

Watermarking techniques utilize statistical methods for verification \cite{kirchenbauer2023}. Since these methods are based on likelihood during both generation and verification, a failure to reproduce likelihood during the verification stage may lead to a decrease in accuracy. On the other hand, robust watermarking techniques against paraphrase attacks have emerged \cite{ren2023}. These methods may exhibit robustness against prompts as well.

\subsection{Towards resilient zero-shot detectors}
Currently, many methods perform likelihood-based detection. Combining these approaches with other methods may lead to more robust detection. One such approach is Intrinsic Dimension \cite{tulchinskii2023}. Intrinsic Dimension refers to the minimum dimension needed to represent a given text. Tulchinskii et al. propose a detector based on Persistent Homology to estimate the Intrinsic Dimension and use it as a score. However, this method requires a constant length of text and was not applicable in our experiment. It would be interesting to explore the application of this method in experiments involving longer texts.

Approaches utilizing representations obtained with masked language models, including Intrinsic Dimension, calculate likelihood in a different way from the detectors used in our experiment, which are based on autoregressive language models. Combining these elements could lead to the development of a more robust zero-shot detector.

\subsection*{Acknowledgement}
This research was supported in part by JSPS international scientific exchanges between Japan and India, Bilateral Program DTS-JSP, grant number JPJSBP120227718, and the Kayamori Foundation of Informational Science Advancement.


\begin{thebibliography}{99}

\bibitem{openai2023}OpenAI. (2023). GPT-4 Technical Report, arXiv e-prints. 

\bibitem{microsoft2023}Microsoft. Microsoft Copilot, Retrieved October 31, 2023, from  \url{https://adoption.microsoft.com/ja-jp/copilot/.}

\bibitem{google2023}Team, G., Anil, R., Borgeaud, S., Wu, Y., Alayrac, J. B., Yu, J., ... \& Ahn, J. (2023). Gemini: a family of highly capable multimodal models. arXiv preprint arXiv:2312.11805.

\bibitem{gehrmann2019}Gehrmann, S., Strobelt, H., \& Rush, A. (2019). GLTR: Statistical Detection and Visualization of Generated Text. In M. R. Costa\-jussà \& E. Alfonseca (Eds.), Proceedings of the 57th Annual Meeting of the Association for Computational Linguistics: System Demonstrations (pp. 111–116). Association for Computational Linguistics. 

\bibitem{mitchell2023} Mitchell, E., Lee, Y., Khazatsky, A., Manning, C. D., \& Finn, C. (2023). DetectGPT: Zero-shot machine-generated text detection using probability curvature. In Proceedings of the 40th International Conference on Machine Learning (ICML'23) (Vol. 202, pp. 24950–24962). JMLR.org

\bibitem{bao2023}Bao, G., Zhao, Y., Teng, Z., Yang, L., \& Zhang, Y. (2023). Fast-DetectGPT: Efficient Zero-Shot Detection of Machine-Generated Text via Conditional Probability Curvature. arXiv preprint arXiv:2310.05130.

\bibitem{su2023} Su, J., Zhuo, T. Y., Wang, D., \& Nakov, P. (2023). DetectLLM: Leveraging Log Rank Information for Zero-Shot Detection of Machine-Generated Text. arXiv preprint arXiv:2306.05540.

\bibitem{hans2024}Hans, A., Schwarzschild, A., Cherepanova, V., Kazemi, H., Saha, A., Goldblum, M., ... \& Goldstein, T. (2024). Spotting LLMs With Binoculars: Zero-Shot Detection of Machine-Generated Text. arXiv preprint arXiv:2401.12070.

\bibitem{liu2024}Liu, S., Liu, X., Wang, Y., Cheng, Z., Li, C., Zhang, Z., ... \& Shen, C. (2024). Does\textsc {DetectGPT} Fully Utilize Perturbation? Selective Perturbation on Model-Based Contrastive Learning Detector would be Better. arXiv preprint arXiv:2402.00263.

\bibitem{sasse2024}Sasse, K., Barham, S., Kayi, E. S., \& Staley, E. W. (2024). To Burst or Not to Burst: Generating and Quantifying Improbable Text. arXiv preprint arXiv:2401.15476.

\bibitem{tulchinskii2023}Tulchinskii, E., Kuznetsov, K., Kushnareva, L., Cherniavskii, D., Barannikov, S., Piontkovskaya, I., ... \& Burnaev, E. (2023). Intrinsic Dimension Estimation for Robust Detection of AI-Generated Texts. arXiv preprint arXiv:2306.04723.



\bibitem{solaiman2019}Solaiman, I., Brundage, M., Clark, J., Askell, A., Herbert-Voss, A., Wu, J., ... \& Wang, J. (2019). Release strategies and the social impacts of language models. arXiv preprint arXiv:1908.09203.

\bibitem{hu2023}Hu, X., Chen, P. Y., \& Ho, T. Y. (2023). RADAR: Robust AI-Text Detection via Adversarial Learning. arXiv preprint arXiv:2307.03838.

\bibitem{dou2024}Dou, Z., Guo, Y., Chang, C. C., Nguyen, H. H., \& Echizen, I. (2024). Enhancing Robustness of LLM-Synthetic Text Detectors for Academic Writing: A Comprehensive Analysis. arXiv preprint arXiv:2401.08046.

\bibitem{liuz2023} Liu, Z., Yao, Z., Li, F., \& Luo, B. (2023). Check Me If You Can: Detecting ChatGPT-Generated Academic Writing using CheckGPT. arXiv preprint arXiv:2306.05524.

\bibitem{kirchenbauer2023} Kirchenbauer, J., Geiping, J., Wen, Y., Katz, J., Miers, I. \& Goldstein, T. (2023). A Watermark for Large Language Models. Proceedings of the 40th International Conference on Machine Learning, in Proceedings of Machine Learning Research 202:17061-17084.

\bibitem{ren2023}Ren, J., Xu, H., Liu, Y., Cui, Y., Wang, S., Yin, D., \& Tang, J. (2023). A Robust Semantics-based Watermark for Large Language Model against Paraphrasing. arXiv preprint arXiv:2311.08721.

\bibitem{shi2023} Shi, Z., Wang, Y., Yin, F., Chen, X., Chang, K. W., \& Hsieh, C. J. (2023). Red Teaming Language Model Detectors with Language Models. arXiv preprint arXiv:2305.19713.

\bibitem{koike2023} Koike, R., Kaneko, M., \& Okazaki, N. (2023). Outfox: Llm-generated essay detection through in-context learning with adversarially generated examples. arXiv preprint arXiv:2307.11729.

\bibitem{lun2023} Lu, N., Liu, S., He, R., \& Tang, K. (2023). Large Language Models can be Guided to Evade AI-Generated Text Detection. arXiv preprint arXiv:2305.10847.

\bibitem{kumarage2023}Kumarage, T., Sheth, P., Moraffah, R., Garland, J., \& Liu, H. (2023). How reliable are ai-generated-text detectors? an assessment framework using evasive soft prompts. arXiv preprint arXiv:2310.05095.

\bibitem{openai_text_classifier2023}OpenAI. (2023). New AI classifier for indicating AI-written text, Retrieved November 30, 2023.

\bibitem{radford2019}Radford, A., Wu, J., Child, R., Luan, D., Amodei, D., \& Sutskever, I. (2019). Language Models are Unsupervised Multitask Learners. Retrieved October 31, 2023, from \url{https://d4mucfpksywv.cloudfront.net/better-language-models/language_models_are_unsupervised_multitask_learners.pdf}

\bibitem{raffel2020}Raffel, C., Shazeer, N., Roberts, A., Lee, K., Narang, S., Matena, M., ... \& Liu, P. J. (2020). Exploring the limits of transfer learning with a unified text-to-text transformer. The Journal of Machine Learning Research, 21(1), 5485-5551.

\bibitem{touvron2023}Touvron, H., Martin, L., Stone, K., Albert, P., Almahairi, A., Babaei, Y., ... \& Scialom, T. (2023). Llama 2: Open foundation and fine-tuned chat models. arXiv preprint arXiv:2307.09288.

\bibitem{almazrouei2023} Almazrouei, E., Alobeidli, H., Alshamsi, A., Cappelli, A., Cojocaru, R., Debbah, M., ... \& Penedo, G. (2023). The falcon series of open language models. arXiv preprint arXiv:2311.16867.

\bibitem{liy2023} Li, Y., Bubeck, S., Eldan, R., Del Giorno, A., Gunasekar, S., \& Lee, Y. T. (2023). Textbooks are all you need ii: phi-1.5 technical report. arXiv preprint arXiv:2309.05463.

\bibitem{narayan2018}Narayan, S., Cohen, S. B., \& Lapata, M. (2018). Don't Give Me the Details, Just the Summary! Topic-Aware Convolutional Neural Networks for Extreme Summarization. In E. Riloff, D. Chiang, J. Hockenmaier, \& J. Tsujii (Eds.), Proceedings of the 2018 Conference on Empirical Methods in Natural Language Processing (pp. 1797–1807). Association for Computational Linguistics

\bibitem{chakraborty2023}Chakraborty, S., Bedi, A. S., Zhu, S., An, B., Manocha, D., \& Huang, F. (2023). On the possibilities of ai-generated text detection. arXiv preprint arXiv:2304.04736.

\bibitem{sadasivan2023}Sadasivan, V. S., Kumar, A., Balasubramanian, S., Wang, W., \& Feizi, S. (2023). Can ai-generated text be reliably detected? arXiv preprint arXiv:2303.11156.

\bibitem{krishna2023}Krishna, K., Song, Y., Karpinska, M., Wieting, J., \& Iyyer, M. (2023). Paraphrasing evades detectors of ai-generated text, but retrieval is an effective defense. arXiv preprint arXiv:2303.13408.

\bibitem{mireshghallah2023}Mireshghallah, F., Mattern, J., Gao, S., Shokri, R., \& Berg-Kirkpatrick, T. (2023). Smaller Language Models are Better Black-box Machine-Generated Text Detectors. arXiv preprint arXiv:2305.09859.

\end{thebibliography}
\end{document}